\documentclass[11pt]{article}
\usepackage[preprint]{acl}
\usepackage{times}
\usepackage{latexsym}
\usepackage[T1]{fontenc}
\usepackage[utf8]{inputenc}
\usepackage{microtype}
\usepackage{booktabs}
\usepackage{amsmath}
\usepackage{graphicx}

\title{Swapping the Judge Changes the Ranking: A Preregistered Study of Cross-Provider Replication in LLM-Judged System Comparisons}

\author{Sangdo Han \\
  CEO Business School \\
  \texttt{sangdohan@ceobizschool.kr} \And
  Moonsoo Kim \\
  CEO Business School \\
  \texttt{mskim@ceobizschool.kr}}

\begin{document}
\maketitle

\begin{abstract}
Practitioners who compare systems with an LLM judge and have no human labels often check a conclusion by repeating it with a judge from another provider. We test how far that check can be trusted. In a preregistered study (with a disclosed post-registration reduction of the judge set), five judge models from three providers re-judged 4,000 Chatbot Arena battles (nine systems, 33 system pairs) that carry human votes, each in both response orders. Agreement with the human vote was low at the verdict level (median Cohen's $\kappa = 0.22$), and system rankings did not survive a change of judge: Kendall's $\tau$ with the human ranking ranged from 0.17 to 0.89 (median 0.67), so the preregistered hypothesis that rankings remain stable was not supported. Conclusions replicated by two low-cost judges from different providers matched the human conclusion more often than single-judge conclusions ($+0.10$ precision), but the gain came from filtering out one poorly aligned judge, and replication involving that judge left a conclusion for only 35\% of comparisons. Judges that agreed with humans reached a stable conclusion with no more verdicts than human votes required; the others failed to converge on 26--70\% of system pairs at the sample sizes tested (up to 100 verdicts per pair). Disagreements with the human vote co-occurred across judges far more often than independence predicts, although a single human vote per battle cannot separate shared judge bias from voter noise.
\end{abstract}

\section{Introduction}

LLM judges are now a routine way to compare language systems \citep{zheng2023judging}. A typical use is a pairwise comparison: a judge model reads one request and two responses and says which response is better, and the verdicts are aggregated into a win rate or a ranking. When human labels are unavailable, the only check on such a conclusion is another judge. A common recommendation is to combine judges from different model families, on the expectation that their individual biases differ \citep{verga2024replacing}.

This paper asks two practical questions about that check. First, when a system-level conclusion obtained with one judge is reproduced by a judge from a different provider, is it more likely to match the conclusion that human votes would give? Second, how many verdicts does a judge need before its system-level conclusion is stable, compared with the number of human votes needed for the same comparison?

A widely repeated observation motivates optimism: judges can disagree with humans on many individual items while still ranking systems similarly \citep{gera-etal-2025-justrank,gao-etal-2025-evaluating}. We preregistered this observation as a replication hypothesis (H1) and three further hypotheses on replication (H2), sample requirements (H3), and co-occurring disagreement (H4). The registered design had six judges; one was dropped after registration, as described in Section~\ref{sec:prereg}. We then re-judged 4,000 battles from a public Chatbot Arena release \citep{chiang2024chatbot,arena140k} with five judge models from three providers. Every battle has a human vote, so verdict-level and system-level quantities are computed on the same items.

The results are mixed, and we report them against the preregistered criteria.
\begin{itemize}
\item \textbf{H1 was not supported.} Verdict-level agreement with humans was low for every judge, and rank correlation with the human ranking depended strongly on the judge ($\tau$ from 0.17 to 0.89; median 0.67 against a preregistered threshold of 0.8).
\item \textbf{H2 was supported, with a narrow mechanism.} Replicated conclusions were more precise than single-judge conclusions, but only because replication removed the wrong conclusions of one poorly aligned judge. Two well-aligned judges gained nothing from replicating each other, and replication with the poorly aligned judge reduced coverage from 100\% to 35\% of comparisons.
\item \textbf{H3 held for four of five judges.} Where a judge converged, it needed no more verdicts than human votes. Three judges failed to converge on a quarter or more of the system pairs within the sizes tested.
\item \textbf{H4 was supported statistically,} but its interpretation is limited because the reference label is a single vote from one user.
\end{itemize}

The study is small (nine systems, 33 system pairs, three low-cost judges in the cross-provider analyses) and its contribution is correspondingly modest: a preregistered measurement, on human-voted battles, of what cross-provider replication buys and what it costs at the level of system-comparison conclusions. We release battle identifiers, all verdicts, the preregistration and deviation records, and the analysis code as ancillary files with this preprint.

\section{Related Work}

\paragraph{Verdict-level versus system-level judge quality.} \citet{gera-etal-2025-justrank} evaluate 48 judge realizations by how well the system ranking they induce matches a human-based leaderboard, and show that verdict-level and system-level quality can diverge. \citet{gao-etal-2025-evaluating} compare combinations of input sets, judges, comparison formats, and aggregation methods and report a similar divergence. \citet{brocai2026pair} show for text ranking that disagreement concentrates on pairs with small quality gaps, which limits its effect on rankings. Our H1 restates the common reading of this line of work as a threshold test and finds that it fails for the judges and data studied here; unlike the leaderboard-based evaluations, our human reference consists of votes on the same battles the judges see.

\paragraph{Judge bias and sample efficiency.} \citet{dorner2025limits} prove that when a judge is no more accurate than the systems it evaluates, bias correction cannot save more than half of the ground-truth labels, and show that judge bias can perturb rankings even when judge accuracy is high. Our H3 analysis is an empirical counterpart on arena data: it compares, pair by pair, the number of judge verdicts and human votes needed to reach the same directional conclusion.

\paragraph{Panels and correlated errors.} Panels of diverse judges were proposed to reduce the bias of any single judge \citep{verga2024replacing}. \citet{kohli2026nine} report that errors of panel members are strongly correlated on NLI items and that the best single judge matches or beats the panel. We observe the same two patterns for pairwise system comparison: disagreements with the human vote co-occur across judges, and a five-judge majority does not beat the best single judge.

\paragraph{Self- and family-preference.} Judges can favor their own generations \citep{panickssery2024llm}, and recent work measures preference at the level of model families \citep{awuni2026who}. A recent analysis by the Arena team had twelve models judge 1,460 English Text Arena battles from 2026 and compared the verdicts with the human votes \citep{arena2026judges}. It reports that most judges favor their own responses and those of sibling models, that judges agree with each other far more often than with the human voter, and that even on the battles where all twelve judges chose the same response the voter agreed only 60.6\% of the time. That analysis is at the verdict level and includes battles in which the judge wrote one of the responses; it does not examine system rankings, replicated system-level conclusions, or sample requirements. Our provider-preference analysis is a prespecified secondary analysis that is consistent with these reports, and we do not present it as a new finding.

\section{Study Design}

\subsection{Data and sample}

We use a public Chatbot Arena release, \texttt{arena-human-preference-140k} \citep{arena140k}, pinned to a fixed revision, which contains 135,634 battles. A battle consists of one user request, the responses of two anonymous systems, and the user's vote. We keep single-turn battles, the first vote of each conversation, votes that choose one of the two responses (ties and ``both bad'' are excluded), and battles whose request and two responses total at most 6,000 tokens; 65,636 battles remain (Appendix~\ref{app:filters}). We do not filter by language. Restricting to English left only three system pairs with at least 100 battles, so the language restriction was dropped before any judge output was produced, and results are reported separately for English and non-English battles.

Systems were added in decreasing order of battle count until system pairs with at least 100 battles could supply 5,000 battles under a common per-pair cap. This rule selected nine systems and 33 system pairs with 111--183 battles each (Appendix~\ref{app:systems}); the comparison graph is connected. The sample was split by pair into a development set (1,000 battles) used for pipeline checks and a confirmation set (4,000 battles) used for all results in this paper. Of the confirmation battles, 2,156 are English, 1,471 are in another identified language, and 373 have no identified language.

\subsection{Judges}

Five judge models from three providers were used: Claude Opus 5.5, Claude Sonnet 5.5, and Claude Haiku 4.5 (Anthropic); GPT-6 Luna (OpenAI); and Gemini 3.1 Flash-Lite (Google). The three low-cost models from different providers (Haiku, Luna, Flash-Lite) form the \emph{economy set} used for the cross-provider analyses, so that provider differences are not confounded with tier differences. The two larger Anthropic models show the effect of tier within one provider. Exact model identifiers and request settings are given in Appendix~\ref{app:protocol}.

\subsection{Judging protocol}

All judges received the same prompt (Appendix~\ref{app:protocol}), which shows the request and the two responses without system names and asks for a brief justification followed by a final line with one of three verdicts: A, B, or tie. Each battle was judged twice with the response order swapped. If the two verdicts selected the same system, that system wins; if they selected different systems or either was a tie, the resolved verdict is a tie. We call a battle order-consistent when the two orders selected the same system or were both ties. A battle with an unparseable or refused response in either order is missing for that judge. For 500 confirmation battles, both orders were judged two additional times to measure run-to-run consistency. Each judge therefore produced 10,000 confirmation responses.

\subsection{Measures}

For pair-level quantities, scores are oriented to the alphabetically first system $X$ of each pair: 1 if $X$ wins, 0 if it loses, and 0.5 for a tie. For a judge or for the human votes, the win rate $w$ of a pair is the mean score, and its \emph{direction} is the sign of $w-0.5$. A direction is \emph{significant} if $|w-0.5| > 1.96\,\mathrm{se}$, where $\mathrm{se}$ is the standard error of the mean. We write $\mathrm{dir}_H$ and $g_H = |w_H - 0.5|$ for the human direction and gap.

\paragraph{Verdict level.} Cohen's $\kappa$ \citep{cohen1960coefficient} between a judge's resolved verdicts and the human votes over the categories \{first system, second system, tie\}, in the original left/right orientation of each battle.

\paragraph{System level.} Bradley--Terry strengths \citep{bradley1952rank} fitted with an MM algorithm \citep{hunter2004mm}, with ties counted as half a win for each side and 0.5 pseudo-wins added in each direction for every observed pair, and Kendall's $\tau_b$ \citep{kendall1938new} and Spearman's $\rho$ between the judge-based and human-based rankings.

\paragraph{Conclusion level.} Let $E$ be the set of pairs whose human direction is significant (23 of 33 pairs in the confirmation set). The single-judge precision $P_1(j)$ is the fraction of pairs in $E$ with a significant direction for judge $j$ whose direction equals $\mathrm{dir}_H$. The replicated precision $P_2(j,k)$ is the same fraction over pairs for which judges $j$ and $k$ are both significant in the same direction. Coverage is the fraction of pairs in $E$ that yield a conclusion.

\paragraph{Sample requirement.} For a source (human votes or a judge), a pair, and $n \in \{25, 50, 100\}$, we draw 1,000 samples of $n$ battles with replacement and record the fraction whose direction equals the full-data $\mathrm{dir}_H$. The requirement $n^*$ is the smallest $n$ for which this fraction is at least 0.95; if no $n$ reaches it, $n^*$ is censored. A size $n$ is skipped for a source when it exceeds that source's number of valid battles for the pair. The preregistered grid also contained $n=200$, which exceeds every pair's confirmation size, and $n=100$ was skipped for the five pairs with fewer than 100 confirmation battles.

\paragraph{Co-occurring disagreement.} Over battles on which all three economy judges give a non-tie verdict, the observed rate $O$ at which all three disagree with the human vote is compared with the product $E_\perp$ of their individual disagreement rates.

\subsection{Hypotheses}
\label{sec:hyp}

\begin{description}
\item[H1] The median $\kappa$ over judges is below 0.6 and the median $\tau$ is above 0.8.
\item[H2] $\Delta = \overline{P_2} - \overline{P_1} > 0$, where the means are over the cross-provider pairs and the members of the economy set.
\item[H3] For pairs with $g_H \ge 0.1$ and a finite human $n^*$, a judge's median ratio $n^*_{\text{judge}}/n^*_{\text{human}}$ is at most 2, with censored values treated as infinite.
\item[H4] $D = O - E_\perp > 0$.
\end{description}
H2 and H4 are tested with one-sided bootstrap tests and Holm correction \citep{holm1979simple}; H1 and H3 are decided by the stated thresholds. Each of the 1,000 bootstrap replicates resamples battles with replacement within pairs, jointly for all judges and the human votes; the set $E$ is held fixed; $p$-values use the plus-one estimator; and precision terms with a zero denominator are excluded from the means and counted.

\subsection{Preregistration and deviations}
\label{sec:prereg}

The prompt, parser, verdict-resolution rule, sample, hypotheses, and analysis code were hashed and frozen before any confirmation response was requested. The set $E$ was fixed from human votes alone. Two changes were made after the freeze.
First, the registered design had six judges, including an open-weight model served by a fourth organization as a fourth economy judge. That judge was dropped for a logistical reason (no additional hosting account was opened), and the decision was taken after the confirmation results of the three Anthropic judges had been seen. H1 is therefore evaluated over five judges and H2 and H4 over three economy judges. Second, after the freeze, the majority-vote rule was changed from the original requirement that all judges have valid verdicts to an abstention rule: ties and missing verdicts abstain, A and B votes are counted, and equal counts (including no votes) give a tie. Both changes were recorded on October 6, 2026, after the Anthropic confirmation results had been seen and before the full five-judge analysis. The majority vote does not enter any hypothesis criterion. No other definition changed. Analyses marked \emph{post hoc} below were not preregistered.

\section{Results}

\subsection{Response quality}

All 50,000 confirmation responses were retrieved. Parse failures were at most four per judge. Opus and Sonnet refused 76 and 39 requests (0.76\% and 0.39\%); the other judges refused none. Resolved verdicts were available for 3,971--4,000 of the 4,000 battles per judge.

\subsection{Verdicts and rankings (H1)}

\begin{table*}[t]
\centering
\small
\begingroup
\setlength{\tabcolsep}{3pt}
\resizebox{\linewidth}{!}{%
\begin{tabular}{lrrrrrrr}
\toprule
Judge & Provider & $\kappa$ & $\tau_b$ & $\rho$ & Order-consistent & Tie & Repeat \\
\midrule
Opus 5.5 & Anthropic & 0.202 & 0.667 & 0.783 & 89.4\% & 11.6\% & 0.868 \\
Sonnet 5.5 & Anthropic & 0.217 & 0.667 & 0.783 & 87.6\% & 13.3\% & 0.845 \\
Haiku 4.5 & Anthropic & 0.266 & 0.889 & 0.967 & 83.0\% & 17.5\% & 0.942 \\
GPT-6 Luna & OpenAI & 0.052 & 0.167 & 0.217 & 78.5\% & 22.1\% & 0.807 \\
Gemini 3.1 Flash-Lite & Google & 0.302 & 0.889 & 0.967 & 85.8\% & 14.9\% & 1.000 \\
Majority of five & -- & 0.267 & 0.778 & 0.900 & -- & -- & -- \\
\bottomrule
\end{tabular}
}
\endgroup

\caption{Agreement with human votes on the confirmation set. $\kappa$: verdict-level agreement; $\tau_b$, $\rho$: correlation between judge-based and human-based Bradley--Terry rankings of nine systems. Order-consistent: battles where the two response orders selected the same system or were both ties. Tie: resolved ties. Repeat: share of repeat battles with three valid resolved verdicts (493--500 per judge) in which all three were identical.}
\label{tab:h1}
\end{table*}

Table~\ref{tab:h1} gives the verdict-level and system-level agreement of each judge with the human votes. The median $\kappa$ is 0.217 (95\% bootstrap interval 0.193--0.240), below the 0.6 threshold. The median $\tau$ is 0.667 (0.556--0.722), below the 0.8 threshold, so \textbf{H1 is not supported}. Two judges reach $\tau = 0.889$, two reach 0.667, and one reaches 0.167. With nine systems, the difference between 0.889 and 0.667 corresponds to four more of the 36 system pairs being ordered differently from the human ranking.

The pattern is similar for English battles ($\tau$ from 0.111 to 0.944) and non-English battles (0.167 to 0.889; Appendix~\ref{app:language}). In a post hoc analysis that leaves out the least aligned judge, the median $\tau$ over the remaining four is 0.778, still below the threshold.

Three further observations qualify these numbers. First, the two low-cost judges Haiku and Flash-Lite agree with human votes more closely than the two larger Anthropic judges in $\kappa$ and in both rank correlations. This says that their preferences are closer to those of Arena voters; it does not show that their verdicts are more correct. Second, the majority vote of the five judges ($\tau = 0.778$) does not beat the best single judges. Third, for every judge the observed $\tau$ lies inside the 95\% range of a random-error model that keeps the judge's tie rate and its agreement rate on non-tie verdicts but places disagreements at random. This does not show that the disagreements are random; it shows that with nine systems the ranking statistic alone cannot tell systematic disagreement from random disagreement. The provider-level analysis in Section~\ref{sec:provider} can.

\begin{figure}[t]
\centering
\includegraphics[width=\columnwidth]{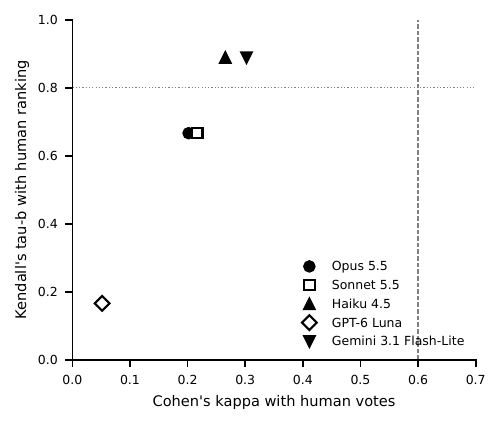}
\caption{Verdict agreement and rank correlation with the human reference. Markers identify individual judges; the reference lines show the preregistered H1 thresholds.}
\label{fig:h1}
\end{figure}

\subsection{Replication across providers (H2)}

\begin{table}[t]
\centering
\small
\begingroup
\setlength{\tabcolsep}{3pt}
\begin{tabular}{p{0.49\linewidth}rr}
\toprule
Source of conclusion & Precision & Coverage \\
\midrule
\multicolumn{3}{l}{\emph{Single judge}} \\
Haiku 4.5 & 22/23 (0.96) & 1.00 \\
GPT-6 Luna & 7/14 (0.50) & 0.61 \\
Gemini 3.1 Flash-Lite & 22/23 (0.96) & 1.00 \\
\midrule
\multicolumn{3}{l}{\emph{Replicated across providers}} \\
Haiku 4.5 +\newline GPT-6 Luna & 7/8 (0.88) & 0.35 \\
Haiku 4.5 +\newline Gemini 3.1 Flash-Lite & 22/23 (0.96) & 1.00 \\
GPT-6 Luna +\newline Gemini 3.1 Flash-Lite & 7/8 (0.88) & 0.35 \\
\midrule
\multicolumn{3}{l}{\emph{Not part of H2}} \\
Opus 5.5 & 15/17 (0.88) & 0.74 \\
Sonnet 5.5 & 17/19 (0.89) & 0.83 \\
Majority of five & 20/21 (0.95) & 0.91 \\
\bottomrule
\end{tabular}
\endgroup

\caption{Precision and coverage of significant directional conclusions on the 23 system pairs whose human direction is significant.}
\label{tab:h2}
\end{table}

Table~\ref{tab:h2} reports precision and coverage on the 23 pairs in $E$. The mean replicated precision is 0.902 and the mean single-judge precision of the economy set is 0.804, giving $\Delta = 0.098$ (95\% interval 0.052--0.194; Holm-adjusted $p = 0.002$). \textbf{H2 is supported.} On all 33 pairs with a nonzero human direction (secondary analysis), $\Delta = 0.125$ (0.028--0.172).

The table also shows where the gain comes from. Haiku and Flash-Lite each reach the human conclusion on 22 of 23 pairs, and the conclusions they share are right on the same 22 of 23: replication between these two judges adds nothing, and the one wrong conclusion is common to both. Luna reaches a significant conclusion on 14 pairs and is right on seven. Requiring agreement with a second judge removes most of Luna's wrong conclusions (seven of eight replicated conclusions are right), but only eight of 23 comparisons then have a conclusion at all. Replication thus acted as a filter against a poorly aligned judge, at a large cost in coverage, and not as a general improvement in validity. A user cannot tell from replication alone which of the two situations applies: agreement between Haiku and Flash-Lite and agreement between Haiku and Luna look the same from the inside. The verdict-level analogue has been reported by \citet{arena2026judges}, where unanimous agreement among twelve judges still matched the human vote on only 60.6\% of battles.

\subsection{Sample requirements (H3)}

\begin{table}[t]
\centering
\small
\begingroup
\setlength{\tabcolsep}{3pt}
\resizebox{\linewidth}{!}{%
\begin{tabular}{lrrrr}
\toprule
Source & $n^*{=}25$ & $50$ & $100$ & Censored \\
\midrule
Human votes & 15 & 7 & 1 & 0 \\
Haiku 4.5 & 22 & 0 & 0 & 1 \\
Gemini 3.1 Flash-Lite & 22 & 0 & 0 & 1 \\
Sonnet 5.5 & 14 & 3 & 0 & 6 \\
Opus 5.5 & 14 & 1 & 0 & 8 \\
GPT-6 Luna & 3 & 3 & 1 & 16 \\
Majority of five & 18 & 2 & 0 & 3 \\
\bottomrule
\end{tabular}
}
\endgroup

\caption{Number of system pairs (of the 23 with human gap $g_H \ge 0.1$) by the smallest sample size at which the source reproduces the human direction in at least 95\% of resamples. Censored: not reached at any size tested.}
\label{tab:h3}
\end{table}

Table~\ref{tab:h3} gives the distribution of $n^*$ over the 23 pairs with $g_H \ge 0.1$; human votes reach a stable direction on all of them. The median ratio of judge to human $n^*$ is 1 for Opus, Sonnet, Haiku, and Flash-Lite and infinite for Luna, so \textbf{H3 holds for four of five judges}. The grid is coarse, and a median ratio of 1 means only that judge and human typically fall in the same grid cell; it is not a statement about every pair.

The censored counts carry most of the information. Haiku and Flash-Lite reproduce the human direction with 25 verdicts on 22 of 23 pairs, while human votes need 50 or more on eight pairs: where these judges agree with the human direction, their verdicts are less noisy than individual votes. Sonnet, Opus, and Luna do not converge on 6, 8, and 16 pairs (26\%, 35\%, and 70\%). For these pairs, the judge's win rate is on the other side of 0.5 or too close to it for the sizes tested, so the available verdicts do not recover the human direction. This matches the theoretical account of \citet{dorner2025limits}, in which judge bias, not judge noise, limits what additional judge labels can deliver.

\begin{figure}[t]
\centering
\includegraphics[width=\columnwidth]{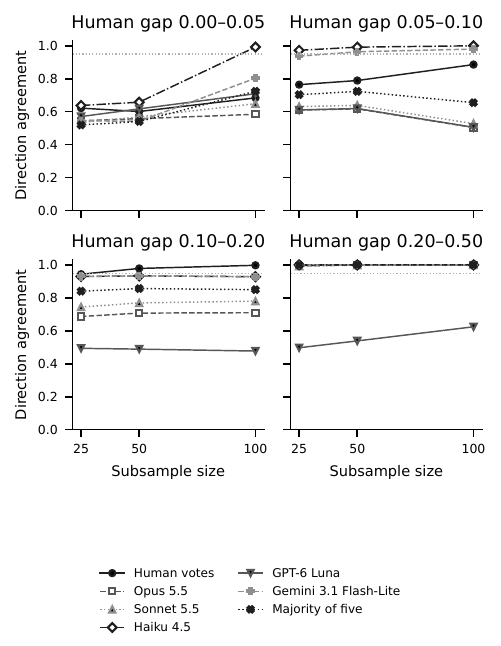}
\caption{Within-pair bootstrap direction agreement, averaged over eligible pairs in each human-gap group. Pair membership can change with sample size because sizes exceeding a source's valid battle count are skipped. The dotted reference is 0.95.}
\label{fig:h3}
\end{figure}

\subsection{Co-occurring disagreement (H4)}

On the 2,351 confirmation battles where all three economy judges give a non-tie verdict, all three disagree with the human vote on 15.8\% of battles, against 3.5\% expected if their disagreements were independent ($D = 0.122$, 95\% interval 0.111--0.133; Holm-adjusted $p = 0.002$). \textbf{H4 is supported.}

This result needs a careful reading. The reference is one vote from one user. If that user preferred the response that most readers would consider worse, every competent judge ``disagrees'' at once, and the disagreements co-occur without any shared judge bias. Our data cannot separate this from shared bias. Two observations are consistent with a substantial contribution of voter noise: at the level of system pairs, where single-vote noise averages out, the economy judges share only one wrong conclusion among the 23 pairs in $E$ (post hoc), and the co-occurrence rate is lower in pairs with larger human gaps (post hoc). We therefore describe H4 as co-occurring disagreement with the human vote and not as evidence of common judge error.

\subsection{Provider preference (prespecified secondary analysis)}
\label{sec:provider}

\begin{table*}[t]
\centering
\small
\begingroup
\setlength{\tabcolsep}{3pt}
\resizebox{\linewidth}{!}{%
\begin{tabular}{lrrrrrr}
\toprule
Judge & Anthropic & Google & OpenAI & Mistral & Alibaba & Own minus others (95\% CI) \\
\midrule
Opus 5.5 & \textbf{+0.101} & +0.081 & -0.019 & -0.159 & -0.190 & +0.173 [0.141, 0.211] \\
Sonnet 5.5 & \textbf{+0.072} & +0.091 & +0.002 & -0.154 & -0.169 & +0.129 [0.098, 0.163] \\
Haiku 4.5 & \textbf{-0.104} & +0.126 & +0.099 & -0.066 & -0.032 & -0.136 [-0.167, -0.105] \\
GPT-6 Luna & +0.245 & -0.114 & \textbf{-0.074} & -0.125 & -0.145 & -0.039 [-0.082, 0.003] \\
Gemini 3.1 Flash-Lite & -0.110 & \textbf{+0.155} & +0.125 & -0.105 & -0.062 & +0.193 [0.161, 0.224] \\
\bottomrule
\end{tabular}
}
\endgroup

\caption{Mean difference between the judge's score and the human score for systems of each provider, over battles in which exactly one of the two systems is from that provider (870--2,163 battles per cell). Positive values mean the judge favors that provider's systems more than human voters do. Bold: the judge's own provider. Last column: own-provider cell minus the mean of the other cells, with 95\% bootstrap interval.}
\label{tab:provider}
\end{table*}

Because the nine selected systems are dominated by three providers, excluding battles that involve a judge's own provider removes 31--63\% of the sample, so we measure preference directly instead. Table~\ref{tab:provider} shows, for each judge and system provider, how much more often the judge favors that provider's systems than human voters do. Opus, Sonnet, and Flash-Lite favor their own provider's systems; Haiku disfavors them; Luna shows no own-provider preference but strongly favors Anthropic systems. All judges rate the Mistral and Alibaba systems lower than human voters do. The preferences are thus systematic and judge-specific, and they are not explained by provider identity alone. For example, human votes rank Claude Opus 4 seventh of nine systems, while the Opus and Sonnet judges rank it fourth. These measurements compare judges with Arena voters and say nothing about which side is right; they are consistent with recent reports of self- and family-preference \citep{panickssery2024llm,awuni2026who,arena2026judges}.

\section{Discussion}

\paragraph{What replication with a second provider can and cannot show.} In this study, agreement between two judges from different providers was informative only in one direction. When a poorly aligned judge was involved, disagreement with a second judge flagged most of its wrong conclusions. But agreement between two judges did not certify a conclusion: the two best-aligned judges shared their one wrong conclusion, and the pair-level precision of replicated conclusions was no higher than that of the better judge alone. Replication is therefore better understood as a screen for an unreliable judge than as a substitute for human validation, and its cost in lost conclusions can be large.

\paragraph{Judge choice matters more than sample size.} For well-aligned judges, 25 verdicts per system pair were enough on almost all pairs with a human gap of at least 0.1. For the other judges, a quarter or more of the pairs never converged. Within the sizes we tested, more verdicts from the same judge did not resolve these pairs; a small human-labeled set that reveals which judge is aligned for the task at hand would address the cause directly.

\paragraph{Stronger models were not better aligned with voters.} The two larger Anthropic judges agreed less with Arena votes than the low-cost judges did. One reading is that larger models apply criteria that differ from those of casual voters, for example by weighting correctness over presentation. Our data do not test this reading.

\section{Limitations}

The study has nine systems and 33 system pairs, and rank correlations over nine systems are coarse. Conclusion-level results rest on 23 pairs, and the H2 effect depends on a single poorly aligned judge. The cross-provider analyses use one low-cost judge per provider; we did not test replication between larger models from different providers. The registered design included a sixth judge, an open-weight economy model, that was dropped after partial results had been seen; although the reason was logistical, this is a deviation from the preregistration. All five judges are proprietary models that may be retired, which limits exact reproduction of the verdicts; we release the verdicts themselves.

The human reference is a single crowd vote per battle from 2025. Such votes are noisy and reflect voter preference, not correctness, and the judged systems are older than most of the judge models. Some battle texts may have been seen by the judges during training. The judging prompt is in English while 37\% of confirmation battles are in another identified language. Ties and ``both bad'' votes were excluded, so the results apply to battles where the voter expressed a preference. We tested one prompt and one verdict-resolution rule; treating order-inconsistent verdicts as ties is one of several reasonable choices. Reasoning settings were not uniform across judges, because providers expose different controls (Table~\ref{tab:execution}): GPT-6 Luna ran with reasoning turned off, while Opus and Sonnet ran with low-effort reasoning. We cannot tell whether Luna's low agreement with human votes reflects the model or this setting, and the H2 effect depends on that judge. The prediction-powered baseline listed in the preregistered plan was not implemented.

\section*{Ethics and Reproducibility}

The data are a public release of user conversations collected between April and July 2025 with pre-existing votes; no new human annotation was collected. The dataset documentation does not describe removal of personal information. We did not apply further filtering; the texts were processed only through the providers' batch interfaces, and we do not redistribute request or response text. The release contains battle identifiers, all judge verdicts with token usage, the frozen analysis code, the preregistration, and the deviation record, from which every number in this paper can be regenerated. The code is released under the MIT License and the verdicts and derived tables under CC BY 4.0; the underlying requests and responses remain under the terms of the original release. Judging cost about USD 157 at the fixed exchange rate used in the cost ledger.

\paragraph{Use of AI assistants.} The judging and analysis pipeline was implemented with an LLM-based coding agent, and LLM assistants were used in planning the analysis and in drafting this manuscript. The authors reviewed the code through its test suite and an independent reproduction run, checked the reported numbers against the released tables, and take responsibility for the content.

\bibliography{refs}

\appendix

\section{Filtering}
\label{app:filters}

\begin{table}[h]
\centering
\small
\begingroup
\setlength{\tabcolsep}{3pt}
\resizebox{\linewidth}{!}{%
\begin{tabular}{lr}
\toprule
Stage & Battles remaining \\
\midrule
Raw release & 135,634 \\
Duplicates removed & 135,619 \\
Two different systems & 135,609 \\
Non-empty responses & 135,472 \\
Single turn & 116,307 \\
First vote per conversation & 92,488 \\
Vote selects one response & 67,617 \\
At most 6,000 tokens & 65,636 \\
\bottomrule
\end{tabular}
}
\endgroup

\caption{Filtering stages. Token counts use the \texttt{cl100k\_base} tokenizer.}
\end{table}

After filtering, 146 system pairs have at least 100 battles. Identical request strings are rare in the sample (39 of 5,000 battles belong to a group of repeated requests) and were kept.

\section{Systems}
\label{app:systems}

The nine systems, in the order of the human-based Bradley--Terry ranking on the confirmation set:
{\small
\begin{enumerate}
\itemsep0pt
\item \texttt{gemini-2.5-pro}
\item \texttt{o3-2025-04-16}
\item \texttt{chatgpt-4o-latest-20250326}
\item \texttt{gemini-2.5-flash}
\item \texttt{qwen3-235b-a22b-no-thinking}
\item \texttt{mistral-medium-2505}
\item \texttt{claude-opus-4-20250514}
\item \texttt{claude-sonnet-4-20250514}
\item \texttt{claude-3-7-sonnet-20250219-}\\\texttt{thinking-32k}
\end{enumerate}}

\begin{table*}[t]
\centering
\small
\begingroup
\setlength{\tabcolsep}{4pt}
\begin{tabular}{p{.12\linewidth}p{.24\linewidth}p{.18\linewidth}p{.21\linewidth}rr}
\toprule
Judge & Model ID & Temperature & Reasoning & Cap & Batch \\
\midrule
Opus 5.5 & claude-\allowbreak opus-\allowbreak 5-\allowbreak 5 & Not sent (unsupported) & adaptive;\newline effort=low & 5120 & Yes \\
Sonnet 5.5 & claude-\allowbreak sonnet-\allowbreak 5-\allowbreak 5 & Not sent (unsupported) & between\_tools;\newline effort=low;\newline no tools & 1024 & Yes \\
Haiku 4.5 & claude-\allowbreak haiku-\allowbreak 4-\allowbreak 5-\allowbreak 20251001 & 0 & disabled & 1024 & Yes \\
GPT-6 Luna & gpt-\allowbreak 6-\allowbreak luna & Not sent (unsupported) & effort=none & 1024 & Yes \\
Gemini 3.1\newline Flash-Lite & gemini-\allowbreak 3.1-\allowbreak flash-\allowbreak lite & 0 & thinkingLevel=minimal & 5120 & Yes \\
\bottomrule
\end{tabular}
\endgroup

\caption{Execution settings for the five judges. Cap is the API output-token limit, including reasoning where applicable. ``Not sent'' denotes an omitted temperature parameter, not a claimed provider default. All requests used batch interfaces.}
\label{tab:execution}
\end{table*}

\section{Judging protocol}
\label{app:protocol}

All requests were sent through the providers' batch interfaces. The target for visible output was 1,024 tokens; for Opus and Flash-Lite the total output limit was set to 5,120 tokens to leave room for reasoning tokens, which does not guarantee a fixed amount of visible output. An initial limit of 300 tokens truncated some responses in a partial development run; the limit was raised and the development run repeated before the freeze. Table~\ref{tab:execution} reports the settings used. Analyses used Python 3.11 with NumPy 2.4.6, SciPy 1.17.1, and pandas 3.0.6; the Bradley--Terry fit, bootstrap, and resampling procedures were implemented directly, and all package versions are pinned in the released lock file.

\paragraph{System prompt.} \emph{You are an impartial evaluator comparing two AI assistant responses to the same user request.}

\paragraph{User prompt.} The request and the two responses are inserted under the headings \emph{[User request]}, \emph{[Response A]}, and \emph{[Response B]}, followed by: \emph{Which response better serves the user's request? Give a brief justification in at most three sentences. Then end with exactly one final line in one of these forms: VERDICT: A / VERDICT: B / VERDICT: TIE}.

After the analysis, all 60,000 development and confirmation responses were downloaded again from the providers and matched to their requests by the server-side request identifier; no mismatch was found (post hoc integrity check).

\section{English and non-English battles}
\label{app:language}

{\centering\makeatletter\def\@captype{table}\makeatother
\small
\begingroup
\setlength{\tabcolsep}{3pt}
\resizebox{\linewidth}{!}{%
\begin{tabular}{lrrrr}
\toprule
Judge & $\kappa$ (EN) & $\tau_b$ (EN) & $\kappa$ (non-EN) & $\tau_b$ (non-EN) \\
\midrule
Opus 5.5 & 0.180 & 0.611 & 0.221 & 0.556 \\
Sonnet 5.5 & 0.212 & 0.667 & 0.219 & 0.667 \\
Haiku 4.5 & 0.264 & 0.944 & 0.284 & 0.889 \\
GPT-6 Luna & 0.047 & 0.111 & 0.045 & 0.167 \\
Gemini 3.1 Flash-Lite & 0.292 & 0.944 & 0.325 & 0.833 \\
Majority of five & 0.258 & 0.722 & 0.281 & 0.778 \\
\bottomrule
\end{tabular}
}
\endgroup

\caption{H1 measures on English (2,156) and non-English (1,471) confirmation battles; 373 battles without an identified language are excluded from this table only.}
\par}

\end{document}